\documentclass[8pt]{article}
\usepackage{spconf,amsmath,graphicx}
\usepackage{epsfig}
\usepackage{subfigure}
\usepackage{algorithm}
\usepackage{algorithmic}
\usepackage{amsthm}
\usepackage{multicol}
\usepackage{multirow}
\usepackage{array}
\usepackage{pslatex}
\usepackage{times,enumerate}
\usepackage[small]{caption}
\usepackage{placeins}
\usepackage[utf8]{inputenc} 
\usepackage[T1]{fontenc}    
\usepackage{url}       
\usepackage{amsfonts}       
\usepackage{amsmath}        
\usepackage{amssymb}
\usepackage{amstext}
\usepackage{amsthm}
\usepackage{graphicx}

\DeclareMathOperator*{\argmin}{arg\,min}

\newcommand{\measurement}{\mathbf{y}}
\newcommand{\signal}{\mathbf{x}}
\newcommand{\reps}{\mathbf{c}} 
\newcommand{\repsHat}{\hat{\reps}} 

\title{Compressed Learning: A Deep Neural Network Approach}
\name{Amir Adler\thanks{This research is supported in part by ERC Grant agreement no. 320649, and in part by the Intel Collaborative Research Institute for Computational Intelligence (ICRI-CI).}, Michael Elad and Michael Zibulevsky}

\address{Computer Science Department, Technion, Haifa 32000, Israel}
\begin{document}
%
\maketitle
\begin{abstract}
Compressed Learning (CL) is a joint signal processing and machine learning framework for inference from a signal, using a small number of measurements obtained by linear projections of the signal. In this paper we present an end-to-end deep learning approach for CL, in which a network composed of fully-connected layers followed by convolutional layers perform the linear sensing and non-linear inference stages. During the training phase, the sensing matrix and the non-linear inference operator are \emph{jointly} optimized, and the proposed approach outperforms state-of-the-art for the task of image classification. For example, at a sensing rate of $1\%$ (only 8 measurements of $28 \times 28$ pixels images), the classification error for the MNIST handwritten digits dataset is $6.46\%$ compared to $41.06\%$ with state-of-the-art.
\end{abstract}

\section{Introduction}
\label{sec:intro}
Compressed learning \cite{Calderbank09compressedlearning} is a mathematical framework that combines compressed sensing \cite{Donoho2006,Candes2008} with machine learning. In contrast to compressed sensing, the goal of CL is inference from the signal rather than signal reconstruction. In the CL framework, the measurement device acquires the signal in the linear projections domain, and the inference is performed in the low-dimensional measurements domain using machine learning tools. CL has diverse applications including compressive image classification \cite{ICIP2016}, reconstruction-free action recognition \cite{7208864,7301371,6467135}, compressive acquisition of dynamic scenes \cite{Sankaranarayanan2010}, compressive least-squares regression \cite{maillard2009compressed}, compressive watermark detection \cite{wang2014compressive}, compressive prediction of protein–protein interactions \cite{zhang2011adaptive}, compressive targets classification \cite{Davenport07thesmashed,Li2013}, and compressive hyperspectral image analysis \cite{hahn2014adaptive}. \\
In this paper we present an end-to-end deep learning \cite{Goodfellow-et-al-2016-Book} solution to the CL problem, and the effectiveness of this approach is demonstrated for the task of image classification \cite{ICIP2016}. The main novelty of the proposed approach is that the sensing matrix is jointly optimized with the inference operator. This is in contrast to previous works, which decouple the sensing matrix choice from the inference operator, and employ standard compressed sensing matrices. \\
The contributions of this paper are two-fold: (1) It presents for the first time, to the best knowledge of the authors, the utilization of a deep neural network for the tasks of compressive linear sensing and non-linear inference; and (2) During training, the proposed network \emph{jointly} optimizes the compressive sensing matrix and the inference operator, leading to a significant advantage compared to state-of-the-art for the task of image classification.\\
This paper is organized as follows: section \ref{Compressed Learning} reviews compressed sensing and learning concepts. Section \ref{The Proposed Approach} presents the end-to-end deep learning approach, and discusses structure and training aspects. Section \ref{Results} evaluates the performance of the proposed approach for compressively classify images, and compares it with state-of-the-art. Section \ref{Conclusions} concludes the paper and discusses future research directions.
\section{Compressed Learning Overview}
\label{Compressed Learning}
\vskip -0.25cm
CL performs inference from compressed sensing measurements, and this section begins by explaining compressed sensing principles, followed by a review of CL concepts.
\subsection{Compressed Sensing}
Given a signal $\signal \in \mathbf{R}^N$, an $M \times N$ sensing matrix $\Phi$ (such that $M \ll N$) and a measurements vector $\measurement = \Phi \signal$, the goal of CS is to recover the signal from its measurements. The sensing rate is defined by $R=M/N$, and since $R \ll 1$ the recovery of $\signal$ is not possible in the general case. According to CS theory \cite{Donoho2006,Candes2008}, signals that have a sparse representation in the domain of some linear transform can be exactly recovered with high probability from their measurements: let $\signal = \Psi \reps $, where $\Psi$ is the inverse transform, and $\reps$ is a sparse coefficients vector with only $S \ll N$ non-zeros entries, then the recovered signal is synthesized by $ \hat{\signal} = \Psi \repsHat$, and $\repsHat$ is obtained by solving the following convex optimization program:
\begin{equation}
 \repsHat  = \argmin_{\reps{'}} \left\|\reps{'}\right\|_1 \text{ subject to } \measurement = \Phi \Psi \reps{'},
\end{equation}
\noindent where $\left\|\alpha\right\|_1$ is the $l_1$-norm, which is a convex relaxation of the $l_0$ pseudo-norm that counts the number of non-zero entries of $\alpha$. The exact recovery of $\signal$ is guaranteed with high probability if $\reps$ is sufficiently sparse and if certain conditions are met by the sensing matrix and the transform.
\subsection{Compressed Learning}

CL was introduced in \cite{Calderbank09compressedlearning}, which showed theoretically that direct inference from compressive measurements is feasible with high classification accuracies. In particular, this work provided analytical results for training a linear Support Vector Machine (SVM) classifier in the compressed sensing domain $\measurement = \Phi \signal$: it proved that under certain conditions the performance of a linear SVM classifier operating in the compressed sensing domain is almost equivalent to the performance of the best linear threshold classifier operating in the signal domain $\signal$. A different approach, termed \emph{smashed filters}, was presented in \cite{Davenport07thesmashed}, in which a generalized maximum-likelihood criterion was employed to design matched filters for target detection in the measurements domain of compressive cameras such as the single-pixel camera \cite{Romberg2008}. This approach was extended and termed \emph{smashed correlation filters} for activity recognition by \cite{7208864}, and for face recognition by \cite{7301371}. A deep learning approach was introduced by \cite{ICIP2016}, in which standard sensing matrices were employed for image classification in the compressed sensing domain. This work utilized convolutional networks that operate on the image domain, and used the following projected measurement vector as the input to the network\footnote{Instead of the true image, and after reshaping it to $\sqrt{N} \times \sqrt{N}$ pixels.}:
\begin{equation}
\label{reprojection_sensing}
\textbf{z}=\Phi^T\measurement \in \mathbf{R}^N.
\end{equation}
By training a network similar to LeNet \cite{Lecun98gradient-basedlearning} for classifying MNIST handwritten digits images, and using the projected measurement $\textbf{z}$ rather than the true image $\signal$, outstanding classification results were obtained by \cite{ICIP2016}, which significantly outperform the smashed filters approach at sensing rates as low as $R=0.01$. This approach was also successfully verified for the challenging task of classifying a subset of the ImageNet dataset, consisting of 1.2 million images and 1,000 categories, again demonstrating excellent classification performance.

\section{The Proposed Approach}
\label{The Proposed Approach}
In this paper we propose an end-to-end deep learning solution for CL, which jointly optimizes the sensing matrix and the inference operator. Our choice is motivated by the outstanding success of convolutional networks for the task of compressive image classification \cite{ICIP2016}, which employed a random sensing matrix (with Gaussian entries) for classifying the MNIST \cite{Lecun98gradient-basedlearning} dataset, and a Hadamard matrix for classifying a subset of the ImageNet dataset. In our approach, the first layer learns and performs the sensing matrix $\widetilde{\Phi}$ stage, and the subsequent layers (a fully-connected layer followed by LeNet layers) perform the non-linear inference stage. Note that the second fully-connected layer performs a similar operator to (\ref{reprojection_sensing}), however, a different matrix $\widetilde{\Psi} \in \mathbf{R}^{N \times M}$ is learned and both the first and the second layers are followed by a ReLU \cite{icml2010_NairH10} activation unit, such that the input $\textbf{z}$ to the first convolutional layer\footnote{The signal $\textbf{z}$ is reshaped to $\sqrt{N} \times \sqrt{N}$ pixels, prior entering the convolutional layer.} is given by:
\begin{equation}
\label{block-based sensing}
\textbf{z} = max(0,\widetilde{\Psi} \mathbf{v}) \in \mathbf{R}^N,\\
\mathbf{v} = max(0,\widetilde{\Phi} \signal).
\end{equation}

\noindent Once the network is trained, the sensing stage (i.e. the first hidden layer) can be detached from the subsequent inference layers, into two separate elements of a CL system.\\
The proposed network architecture includes the following layers and elements: (1) an input layer with $N$ nodes; (2) a compressed sensing fully-connected layer with $NR$ nodes, $R\ll1$ (its weights form the sensing matrix); (3) ReLU activation units; (4) a fully-connected layer that expands the output of the sensing layer to the original image dimensions $N$; (5) ReLU activation units; (6) a convolution layer with kernel sizes of $5 \times 5$, which generates 6 feature map; (7) ReLU activation units; (8) maxpooling layer which selects the maximum of $2 \times 2$ feature maps elements, with a stride of 2 in each dimension; (9) a convolution layer with kernel sizes of $5 \times 5$, which generates 16 feature map; (10) ReLU activation units; (11) maxpooling layer which selects the maximum of $2 \times 2$ feature maps elements, with a stride of 2 in each dimension; (12) reshape operator that reshapes the 16 $4 \times 4$ max-pooled features maps into a single 256-dimensional vector; (13) a fully connected layer of 256 to 120 nodes; (14) ReLU activation units; (15) a fully connected layer of 120 to 84 nodes; (16) ReLU activation units; and (17) a softmax layer with 10 outputs (corresponding to the 10 MNIST classes).
\begin{table*}[]
  \caption{Classification Error ($\%$) for the MNIST handwritten digits dataset vs. sensing rate $R=M/N$ (averaged over 10,000 test images):}
  \label{Reconstruction-Quality}
  \centering
  \begin{tabular}{ccccc}
    \hline
    Sensing Rate & No. of Measurements & Smashed Filters \cite{Davenport07thesmashed} & Random Sensing + CNN \cite{ICIP2016} & Proposed \\
    \hline
    0.25         &     196             &  27.42$\%$                                   & 1.63$\%$              &  \textbf{1.56$\%$}  \\
    0.1          &     78              &  43.55$\%$                                   & 2.99$\%$              &  \textbf{1.91$\%$}  \\
    0.05         &     39              &  53.21$\%$                                   & 5.18$\%$              &  \textbf{2.86$\%$}  \\
    0.01         &     8               &  63.03$\%$                                   & 41.06$\%$             & \textbf{ 6.46$\%$}  \\

    \hline
  \end{tabular}
\end{table*}

\section{Performance Evaluation}
\label{Results}
This section provides performance evaluation results of the proposed approach\footnote{A software package implementing the proposed approach is available at: \url{http://www.cs.technion.ac.il/~adleram/CL_DNN_2016.zip}} vs. the Smashed Filters \cite{Davenport07thesmashed} approach and random sensing matrix followed by convolutional network \cite{ICIP2016} approach. We have trained\footnote{The network was implemented using Torch7 \cite{Collobert_NIPSWORKSHOP_2011} scripting language, and trained on NVIDIA Titan X GPU card.} the proposed network from the training images of the MNIST dataset, using the Stochastic Gradient Descent algorithm with a learning rate of $0.0025$ and 100 epochs. Classification error performance was evaluated for sensing rates in the range of $R=0.01$ to $R=0.25$, and averaged over the collection of 10,000 MNIST test images. Classification error results are summarized in Table \ref{Reconstruction-Quality}, and reveal a consistent advantage of the proposed approach, which increases significantly for lower sensing rates.

\vskip -0.5cm
\section{Conclusions}
\label{Conclusions}
This paper presents an end-to-end deep neural network approach to compressed learning, in which the sensing matrix and the non-linear inference operator are jointly optimized during the training phase. This is demonstrated to outperform state-of-the-art deep learning approach, which does not optimize the sensing matrix, and uses standard sensing matrices.
\vskip -0.25cm
\bibliographystyle{IEEEbib}
\bibliography{SPL2017_Bibliography}

\end{document}